\documentclass[conference]{IEEEtran}
\IEEEoverridecommandlockouts
\usepackage{cite}
\usepackage{amsmath,amssymb,amsfonts}
\usepackage{algorithmic}
\usepackage{graphicx}
\usepackage{textcomp}
\usepackage{xcolor}

\usepackage{booktabs}
\usepackage{lineno}
\usepackage{multirow}
\usepackage{multicol}
\usepackage{mathrsfs}
\usepackage{threeparttable}

\def\BibTeX{{\rm B\kern-.05em{\sc i\kern-.025em b}\kern-.08em
    T\kern-.1667em\lower.7ex\hbox{E}\kern-.125emX}}
\begin{document}

\title{Boosting Semi-Supervised Medical Image Segmentation via Masked Image Consistency and Discrepancy Learning}


\author{
\IEEEauthorblockN{
Pengcheng Zhou\textsuperscript{1}\textsuperscript{\dag},
Lantian Zhang\textsuperscript{3}\textsuperscript{\dag},
Wei Li\textsuperscript{2}\IEEEauthorrefmark{1}
\thanks{\textsuperscript{\dag} Equal Contribution. \IEEEauthorrefmark{1} Corresponding Author.}
}
\\
\IEEEauthorblockA{\textsuperscript{1}School of Information and Communication Engineering, Beijing University of Posts and Telecommunications, Beijing, China}

\IEEEauthorblockA{\textsuperscript{2}School of Artificial Intelligence, Beijing University of Posts and Telecommunications, Beijing, China}

\IEEEauthorblockA{\textsuperscript{3}School of Computer Science and Engineering, Southeast University, Nanjing, China\\leesoon@bupt.edu.cn}
}
\maketitle
\begin{abstract}
Semi-supervised learning is of great significance in medical image segmentation by exploiting unlabeled data. Among its strategies, the co-training framework is prominent. However, previous co-training studies predominantly concentrate on network initialization variances and pseudo-label generation, while overlooking the equilibrium between information interchange and model diversity preservation. In this paper, we propose the Masked Image Consistency and Discrepancy Learning (MICD) framework with three key modules. The Masked Cross Pseudo Consistency (MCPC) module enriches context perception and small sample learning via pseudo-labeling across masked-input branches. The Cross Feature Consistency (CFC) module fortifies information exchange and model robustness by ensuring decoder feature consistency. The Cross Model Discrepancy (CMD) module utilizes EMA teacher networks to oversee outputs and preserve branch diversity. Together, these modules address existing limitations by focusing on fine-grained local information and maintaining diversity in a heterogeneous framework. Experiments on two public medical image datasets, AMOS and Synapse, demonstrate that our approach outperforms state-of-the-art methods.
\end{abstract}

\begin{IEEEkeywords}
Medical Image Segmentation, Semi-supervised Learning, Heterogeneous Framework, Cross Feature Consistency
\end{IEEEkeywords}

\section{Introduction}
\label{sec:intro}
Medical image segmentation is vital for computer-aided diagnosis as precise anatomical structure delineation is crucial ~\cite{pan2023semi}. Conventionally, a large amount of high-quality labeled data is needed for effective segmentation. However, annotating medical image data, especially 3D volumes, is highly labor-intensive and time-consuming~\cite{zhao2023rcps}. Hence, training methods with small labeled sets, like semi-supervised learning (SSL), have been actively explored. SSL uses a large unlabeled set and a small labeled set for model training. Among SSL approaches, consistency learning is particularly effective, aiming to minimize the differences in model responses from diverse views of unlabeled data~\cite{ouali2020semi,chen2021semi,cao2024icr}. 

Consistency learning methods obtain different views through data augmentation \cite{berthelot2019mixmatch} or outputs of differently initialized networks \cite{tarvainen2017mean,chen2021semi,ke2020guided}. The mean teacher (MT) method \cite{tarvainen2017mean,wang2021tripled,liu2022perturbed} combines perturbations and averages network parameters to generate pseudo-labels for unlabeled data. Schemes like uncertainty-guided threshold \cite{yu2019uncertainty,hang2020local} and multi-task assistance \cite{wang2021tripled,luo2021semi} enhance teacher-student method generalization. However, the teacher-student scheme's domain-specific transfer \cite{berthelot2019mixmatch} may cause issues like convergence to similar local minima and confirmation bias, leading to the co-training framework. In co-training, two differently initialized models mutually supervise via generating pseudo-labels for unlabeled data during training, having a lower chance of converging to the same local minima than the teacher-student model. Recent studies show co-training achieves effective consistency regularization via cross-supervision between two independent networks. In Cross pseudo supervision (CPS) \cite{chen2021semi}, cross supervision uses two identically structured but differently initialized networks. Lin et al. \cite{lin2022cld} proposed CLD to handle data bias but failed due to CPS overfitting. Wang et al. \cite{wang2023dhc} proposed the DHC framework to address biases and overcome CPS \cite{chen2021semi} baseline drawbacks. However, prior works in the co-training framework have mainly focused on network initialization differences and pseudo-label generation, neglecting the balance between information exchange and model diversity preservation. For example, they lack effective means to ensure models can explore different feature perspectives while maintaining individuality, which may result in suboptimal feature extraction and limited generalization in complex medical image data processing.

In this work, we propose the Masked Image Consistency and Discrepancy Learning (MICD) framework with three key modules. The Masked Cross Pseudo Consistency (MCPC) module utilizes pseudo-labeling across two branches where one branch's original input-derived pseudo-labels supervise the other's masked input learning. This is designed to enhance context understanding and small sample perception as masking forces the model to focus on different regions, promoting more accurate pixel-level predictions. The Cross Feature Consistency (CFC) module ensures that, despite input perturbations, the decoder features of both branches remain consistent. This strengthens information exchange and model robustness, a crucial aspect of the heterogeneous dual-branch structure as it allows for better utilization of the complementary nature of the branches. The Cross Model Discrepancy (CMD) module employs EMA teacher networks for each branch to supervise their respective outputs. Overall, the MCPC lays the groundwork for learning, the CFC fortifies the information transmission and stability, and the CMD ensures the diversity of branches. This design is to maintain the individuality and learning diversity of the branches. They effectively embody the heterogeneous dual-branch concept described in ~\cite{krogh1994neural}. It prevents premature coupling and ensures branches continuously learn distinct valuable information, jointly driving the framework to achieve high segmentation performance and address the class imbalance issue. 

\begin{figure*}[!ht]
    \centering
    \includegraphics[scale=0.95]{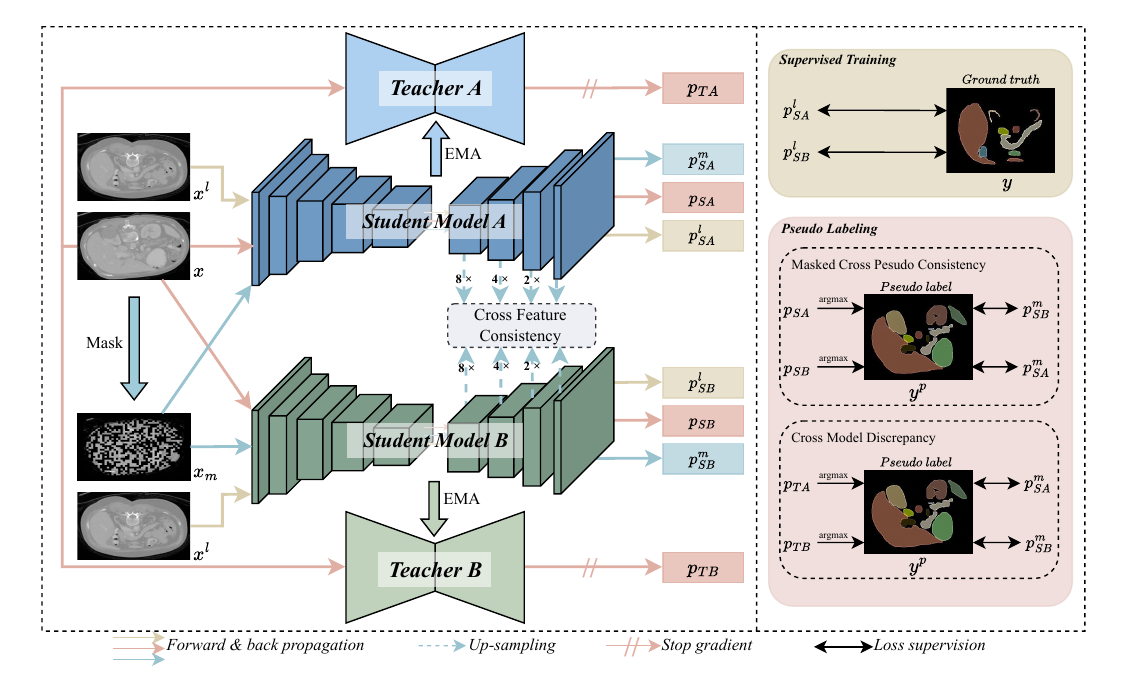}
    \vspace{-0.75cm}
    \caption{An overview of the proposed Masked Image Consistency and Discrepancy Learning (MICD) architecture. Our MICD framework consists of two VNet models and their Teacher networks smoothed via EMA, allowing the peer networks to complement each other through various training strategies. These strategies include Masked Cross Pseudo Consistency (MCPC), Cross Feature Consistency (CFC), and Cross Model Discrepancy (CMD).}
    \label{fig:framework}
\end{figure*}

In summary, the main contributions are:
\begin{itemize}
    \item A novel MICD semi-supervised learning framework that addresses existing limitations by focusing on fine-grained local information and maintaining diversity in a heterogeneous framework.
    \item We propose MCPC to capture fine-grained local semantics and emphasize minority classes, while using CFC to ensure stability and enhance information exchange, and ultimately employing CMD with EMA to prevent premature coupling, thereby ensuring robust learning.
    \item Experiments on two public medical image datasets, AMOS~\cite{amos} and Synapse~\cite{synapse}, demonstrate the superior performance of our approach compared to state-of-the-art (SOTA) methods.
\end{itemize}


\section{Methods}
\subsection{Overview}
The overall framework of the proposed Masked Image Consistency and Discrepancy Learning (MICD) is depicted in Fig.~\ref{fig:framework}. It encompasses three core modules, namely Masked Cross Pseudo Consistency (MCPC), Cross Feature Consistency (CFC), and Cross Model Discrepancy (CMD).

For the semi-supervised medical image segmentation task, two datasets are involved. There is a labeled set $D^{l}=\{(x_i^{l},y_i^{l})\}_{i = 1}^{N}$, where $N$ represents the number of annotated images, and $y_i^{l}$ is the label corresponding to $x_i^{l}$. Additionally, an unlabeled set $D^{u}=\{x_j^{u}\}_{j = 1}^{M}$ exists, which consists of $M$ raw images. The objective of semi-supervised methods is to enhance the segmentation ability by making full use of both $D^{l}$ and $D^{u}$.

Firstly, we initialize two student models, denoted as $A_S: f(x;\theta_A)$ and $B_S: f(x;\theta_B)$, which are derived from VNet~\cite{milletari2016v} . We adopt the strategy proposed in ~\cite{wang2023dhc}  that incorporates two distinct weighting approaches. These weights are respectively denoted as $W^{diff}$ and $W^{dist}$ to initialize the training differences between the two models.

Moreover, two teacher models $A_T: f(x;\theta'_A)$ and $B_T: f(x;\theta'_B)$ are generated from $A_S$ and $B_S$ respectively through the exponential moving average (EMA) of the weights $\theta$ of the student model. The teacher parameters $\theta'_t$ are updated as follows:
\begin{equation}
\theta'_t = \alpha\theta'_t + (1 - \alpha)\theta_t
\end{equation}
where $t$ represents the training iteration, and $\alpha$ is a hyper-parameter that controls the updating rate.

Furthermore, we define masks to introduce stochastic perturbations. The masks are generated by specifying the mask ratio $r$ (indicating the percentage of the image that should be masked) and the mask patch size $s$ (e.g., $3\times3\times3$). In this paper, all mask operations utilize the defined mask operator, which is expressed as:
\begin{equation}
x^{m} = \mathcal{M}(r, s) \odot x
\end{equation}

\subsection{Masked Cross Pseudo Consistency}
To facilitate the acquisition of fine-grained local and contextual information, and to enhance the diversity of the model through data augmentation, we introduce a Masked Cross Pseudo Consistency (MCPC) module. 


For an input image $x_i \in D^{l}\cup  D^{u}$, we apply a masking operator $\mathcal{M}$ to obtain the masked image $x_{i}^{m} = \mathcal{M} \odot x_{i}$. This masked image is then processed by two student models, resulting in probabilities:
$p_{i}^{mA_{S}}=f(x_{i}^{m};\theta_{A})$, $p_{i}^{mB_{S}}=f(x_{i}^{m};\theta_{B})$. From these probabilities, we derive pseudo-labels by selecting the class with the highest probability: $\hat{y}_{i}^{mA_{S}} = \arg\max{p_{i}^{mA_{S}}}$, $\hat{y}_{i}^{mB_{S}} = \arg\max{p_{i}^{mB_{S}}}$. The masked cross pseudo consistency loss is then defined to enforce consistency between the two student models' predictions using these pseudo-labels:
\begin{equation}
\begin{split}
\mathcal{L}_{cps} =\frac{1}{N+M} \sum_{i=1}^{N+M} \Big( &  W_i^{diff} \mathcal{L}_{ce}(p_{i}^{A_S}, \hat{y}_{i}^{mB_S}) + \\& W_i^{dist} \mathcal{L}_{ce}(p_{i}^{B_S}, \hat{y}_{i}^{mA_S}) \Big)
\end{split}
\end{equation}

\subsection{Cross Feature Consistency}

In co-training models, pseudo supervision alone proves insufficient as considerable information may be lost during supervision. To enhance pseudo supervision consistency, we introduce the Cross Feature Consistency (CFC) mechanism.

For an input image $x_i \in D^{l}\cup D^{u}$, two student models, $A_S$ and $B_S$ with parameters $\theta_{A}$ and $\theta_{B}$ respectively, extract features from four layers of the decoder. Specifically, $A_S$ generates features $d_{1i}^{A_{S}}, d_{2i}^{A_{S}}, d_{3i}^{A_{S}}, d_{4i}^{A_{S}} = f(x_i;\theta_{A})$, and $B_S$ yields features $d_{1i}^{B_{S}}, d_{2i}^{B_{S}}, d_{3i}^{B_{S}}, d_{4i}^{B_{S}} = f(x_i;\theta_{B})$. To measure and enforce the consistency between these feature representations, we define the cross feature consistency loss as follows:

\begin{equation}
\mathcal{L}_{con} = \frac{1}{N + M} \sum_{i = 1}^{N + M} \sum_{k = 1}^{4} \Big( \lambda_{k} \mathcal{L}_{kl} (d_{k,i}^{A_{S}}, d_{k,i}^{B_{S}}) \Big)
\end{equation}

Here, $\lambda_{k}$ is a hyper-parameter set to $\lambda_{k} = 0.2\times{k}$ for $k = 1,2,3,4$. This setting is based on the fact that deeper decoder layers ($k$ increasing) capture more abstract and semantically rich features critical for accurate segmentation. By assigning greater weights to these layers, we emphasize their consistency. $\mathcal{L}_{kl}$ represents the KL loss, which quantifies the divergence between the feature representations of the two models at each layer. Minimizing $\mathcal{L}_{con}$ during training encourages the two student models to learn consistent feature representations across different decoder layers, thus improving segmentation performance within the CPS paradigm. 

\subsection{Cross Model Discrepancy}
In co-training models, the heterogeneity or discrepancy among them is significant as greater diversity usually leads to better performance outcomes in semi-supervised learning tasks. To leverage this discrepancy, we adopt the commonly used teacher-student strategy to maintain and enhance the diversity of the two-branch models in our proposed framework. 

Given an input image $x_i$ where $x_i \in D^{l}\cup D^{u}$, we first consider two teacher models, $A_T$ and $B_T$. These teacher models process $x_i$ to compute probabilities. Specifically, $A_T$ computes $p_{i}^{A_{T}} = f(x_i;\theta'_{A})$, and $B_T$ computes $p_{i}^{B_{T}} = f(x_i;\theta'_{B})$. From these probabilities, we derive the pseudo-labels. For model $A_T$, the pseudo-label $\hat{y}_{i}^{A_{T}}$ is obtained by performing the argmax operation on $p_{i}^{A_{T}}$, that is $\hat{y}_{i}^{A_{T}} = \arg\max{p_{i}^{A_{T}}}$. Similarly, for model $B_T$, the pseudo-label $\hat{y}_{i}^{B_{T}}$ is derived as $\hat{y}_{i}^{B_{T}} = \arg\max{p_{i}^{B_{T}}}$.

Next, we apply a mask operator $\mathcal{M}$ to $x_i$, resulting in a masked version of the input image denoted as $x_{i}^{m}$, which is calculated as $x_{i}^{m} = \mathcal{M} \odot x_{i}$. Subsequently, two student models, $A_S$ and $B_S$, process this masked input image. $A_S$ computes the probability $p_{i}^{mA_{S}} = f(x_{i}^{m};\theta_{A})$, from which we obtain the corresponding pseudo-label $\hat{y}_{i}^{mA_{S}} = \arg\max{p_{i}^{mA_{S}}}$. In the same way, $B_S$ computes $p_{i}^{mB_{S}} = f(x_{i}^{m};\theta_{B})$, and its associated pseudo-label is $\hat{y}_{i}^{mB_{S}} = \arg\max{p_{i}^{mB_{S}}}$.

To quantify and optimize the discrepancy between the student and teacher models, we define the discrepancy loss, which is formulated as follows:

\begin{equation}
\begin{split}
\mathcal{L}_{{dis}} = \frac{1}{N + M} \sum_{i = 1}^{N + M} \Big( \mathcal{L}_{{ce}}(p_{i}^{mA_S}, \hat{y}_{i}^{A_T}) + \mathcal{L}_{{ce}}(p_{i}^{mB_S}, \hat{y}_{i}^{B_T}) \Big)
\end{split}
\end{equation}

Here, $\mathcal{L}_{ce}$ represents the cross entropy loss, which is a widely used metric in machine learning for evaluating the difference between predicted probabilities and actual labels (in this case, pseudo-labels). Minimizing the discrepancy loss during training enables student models to learn while maintaining a certain difference from teacher models under their guidance, thus preserving the diversity of the two branches. 

\begin{table*}[t]
\scriptsize
\caption{Quantitative comparison between MICD and SOTA SSL segmentation methods on \textbf{5\% labeled AMOS dataset}. ‘General’ or ‘Imbalance’ indicates whether the methods consider the class imbalance issue or not. \textcolor{red}{The best results are shown in red}, and \textcolor{orange}{the second-best results are shown in orange}. 
}
\label{sota1}
\resizebox*{\linewidth}{!}{

\begin{tabular}{c|c|c@{\ \ }c|ccccccccccccccc}
\toprule
\multicolumn{2}{c|}{\multirow{2}{*}{Methods}}   &Avg. &Avg. &\multicolumn{15}{c}{Average Dice of Each Class}    \\ 
\multicolumn{2}{c|}{}       &Dice &ASD &Sp &RK &LK &Ga &Es &Li &St &Ao &IVC &PA &RAG &LAG &Du &Bl &P/U \\
\midrule
\multirow{10}{*}{\rotatebox{90}{General}}
&V-Net (fully)      &76.50 &2.01 &92.2 &92.2	&93.3 &65.5	&70.3 &95.3	&82.4 &91.4	&85.0 &74.9	&58.6 &58.1	&65.6 &64.4	&58.3   \\ 
\midrule
& UA-MT~\cite{yu2019uamt}$^\dagger$     &42.16 &15.48 &59.8 &64.9 &64.0 &35.3 &34.1 &77.7 &37.8 &61.0 &46.0 &33.3 &26.9 &12.3 &18.1 & 29.7 &31.6    \\
& URPC~\cite{luo2021urpc}$^\dagger$     &44.93 &27.44 &67.0 &64.2 &67.2 &36.1 &0.0 &83.1 &45.5 &67.4 &54.4 &\textcolor{orange}{46.7} &0.0 &\textcolor{red}{\textbf{29.4}} &\textcolor{orange}{35.2} &44.5 &33.2  \\
& CPS~\cite{chen2021cps}$^\dagger$      &41.08 &20.37 &56.1 &60.3 &59.4 &33.3 &25.4 &73.8 &32.4 &65.7 &52.1 &31.1 &25.5 &6.2 &18.4 &40.7 &35.8 \\
& SS-Net~\cite{wu2022ssnet}$^\dagger$       &33.88 &54.72 &65.4 &68.3 &69.9 &37.8 &0.0 &75.1 &33.2 &68.0 &56.6 &33.5 &0.0 &0.0	&0.0 &0.2 &0.2  \\
& DST~\cite{chen2022dst}$^\star$        &41.44 &21.12 &58.9 &63.3 &63.8 &37.7 &29.6 &74.6 &36.1 &66.1 &49.9 &32.8 &13.5 &5.5 &17.6 &39.1 &33.1 \\
& DePL~\cite{wang2022depl}$^\star$      &41.97 &20.42 &55.7 &62.4 &57.7 &36.6 &31.3 &68.4 &33.9 &65.6 &51.9 &30.2 &23.3 &10.2 &20.9 &43.9 &\textcolor{red}{\textbf{37.7}}    \\ 
\midrule
\multirow{7.5}{*}{\rotatebox{90}{Imbalance}} 
& Adsh~\cite{guo2022adsh}$^\star$       &40.33 &24.53 &56.0	&63.6 &57.3	&34.7 &25.7	&73.9 &30.7	&65.7 &51.9	&27.1 &20.2	&0.0 &18.6 &43.5 &35.9  \\ 
& CReST~\cite{wei2021crest}$^\star$     &46.55 &14.62 &66.5 &64.2 &65.4 &36.0 &32.2	&77.8 &43.6	&68.5 &52.9 &40.3 &24.7 &19.5 &26.5	&43.9 &36.4  \\
& SimiS~\cite{simis}$^\star$        &47.27 &\textcolor{orange}{11.51} &\textcolor{orange}{77.4} &\textcolor{orange}{72.5} &68.7 &32.1 &14.7 &\textcolor{red}{\textbf{86.6}} &\textcolor{orange}{46.3} &\textcolor{orange}{74.6}	&54.2 &41.6	&24.4 &17.9	&21.9 &\textcolor{orange}{47.9} &28.2  \\
& Basak \textit{et al.}~\cite{basak2022addressing}$^\dagger$        &38.73 &31.76 &68.8	&59.0 &54.2	&29.0 &0.0	&83.7 &39.3	&61.7 &52.1	&34.6 &0.0	&0.0 &26.8 &45.7 &26.2 \\
& CLD~\cite{lin2022cld}$^\dagger$       &46.10 &15.86 &67.2	&68.5 &\textcolor{orange}{71.4} &41.0 &21.0 &76.1 &42.4 &69.8 &52.1 &37.9 &24.7 &23.4	&22.7 &38.1	&35.2    \\
& DHC~\cite{wang2023dhc}       &\textcolor{orange}{49.53}	&13.89 &68.1 &69.6 &71.1 &\textcolor{orange}{42.3} &\textcolor{orange}{37.0}	&76.8 &43.8	&70.8 &\textcolor{orange}{57.4} &43.2 &\textcolor{orange}{27.0} &\textcolor{orange}{28.7} &29.1 &41.4 &\textcolor{orange}{36.7} \\ 
&\textbf{MICD (ours)}        &\textcolor{red}{\textbf{56.11}} &\textcolor{red}{\textbf{6.22}}	&\textcolor{red}{\textbf{78.2}} &\textcolor{red}{\textbf{77.3}}	&\textcolor{red}{\textbf{75.3}} &\textcolor{red}{\textbf{46.5}} &\textcolor{red}{\textbf{52.9}}	&\textcolor{orange}{84.9} &\textcolor{red}{\textbf{53.7}} &\textcolor{red}{\textbf{82.7}} &\textcolor{red}{\textbf{65.6}} &\textcolor{red}{\textbf{51.5}} &\textcolor{red}{\textbf{32.9}} &23.4 &\textcolor{red}{\textbf{36.9}} &\textcolor{red}{\textbf{49.8}} &30.0   \\ 

\bottomrule
\end{tabular}
}
\begin{threeparttable}
 \begin{tablenotes}
        \scriptsize
        \item[$\dagger$] implement semi-supervised segmentation methods on our dataset.
        \item[$\star$] extend semi-supervised classification methods to segmentation with CPS as the baseline.
\end{tablenotes}
\end{threeparttable}

\end{table*}

\subsection{Overall Training Objective}
For an input image $x_i$ and its corresponding true label $y_i$ from the labeled dataset $D^{l}$, models $A_S$ and $B_S$ compute output probability maps as $p_{i}^{A_S} = f(x_i;\theta_{A})$ and $p_{i}^{B_S} = f(x_i;\theta_{B})$. The supervised loss $\mathcal{L}_{sup}$ is defined as:

\begin{equation}
\mathcal{L}_{sup} = \frac{1}{N} \sum_{i = 1}^{N} \left(W_i^{diff} \mathcal{L}_{s}(p_i^{A_S}, y_i) + W_i^{dist} \mathcal{L}_{s}(p_i^{B_S}, y_i) \right) 
\end{equation}

where $\mathcal{L}_{s}$ combines the cross entropy loss $\mathcal{L}_{ce}$ and the Dice loss $\mathcal{L}_{dice}$ for a comprehensive evaluation. Weights $W_i^{diff}$ and $W_i^{dist}$ balance contributions from $A_S$ and $B_S$.

Combining $\mathcal{L}_{sup}$ with losses from other modules gives the final training objective $\mathcal{L}_{total}$ for the MICD framework:

\begin{equation}
\mathcal{L}_{total} = \mathcal{L}_{sup} + \beta (\mathcal{L}_{cps} + \mathcal{L}_{con} + \mathcal{L}_{dis})
\end{equation}

Here, $\beta$ is a trade-off weight following an epoch-dependent Gaussian ramp-up strategy \cite{lin2022cld} to adaptively adjust loss importance during training, ensuring effective convergence and good performance. 
\begin{table}[t] 
\centering
\setlength{\tabcolsep}{12pt}
\scriptsize
\caption{Quantitative comparison between MICD and SOTA SSL segmentation methods on \textbf{2\% and 10\% labeled AMOS dataset}. \textcolor{red}{The best results are shown in red}, and \textcolor{orange}{the second-best results are shown in orange}. 
}
\label{sota3}
\resizebox{\columnwidth}{!}{ 

\begin{tabular}{c|c|c@{\ \ }c|c@{\ \ }c}
\toprule
\multicolumn{2}{c|}{}   &\multicolumn{2}{c|}{2\%}    &\multicolumn{2}{c}{10\%} \\
\multicolumn{2}{c|}{\multirow{1.5}{*}{Methods}}  &Avg. &Avg. &Avg. &Avg.    \\ 
\multicolumn{2}{c|}{}   &Dice &ASD &Dice &ASD   \\
\midrule
\multirow{10}{*}{\rotatebox{90}{General}}
& V-Net (fully) &76.50 &2.01 &76.50 &2.01 \\ 
\midrule
& UA-MT~\cite{yu2019uamt}$^\dagger$     &33.96 &22.43 &40.60 &38.45 \\
& URPC~\cite{luo2021urpc}$^\dagger$     &38.39 &37.58 &49.09 &29.69 \\
& CPS~\cite{chen2021cps}$^\dagger$      &31.78 &39.23 &54.51 &7.84 \\
& SS-Net~\cite{wu2022ssnet}$^\dagger$   &17.47 &59.05 &38.91 &53.43 \\
& DST~\cite{chen2022dst}$^\star$        &31.94 &39.15 &52.24 &17.66 \\
& DePL~\cite{wang2022depl}$^\star$      &31.56 &40.70 &56.76 &6.70 \\ 
\midrule
\multirow{7.5}{*}{\rotatebox{90}{Imbalance}}
& Adsh~\cite{guo2022adsh}$^\star$       &30.30 &42.48 &54.92 &8.07 \\  
& CReST~\cite{wei2021crest}$^\star$     &34.13 &20.15 &60.74 &4.65 \\  
& SimiS~\cite{simis}$^\star$            &36.89 &26.16 &57.48 &4.46 \\  
& Basak \textit{et al.}~\cite{basak2022addressing}$^\dagger$        &29.87 &35.55 &53.66 &8.50  \\      
& CLD~\cite{lin2022cld}$^\dagger$       &36.23 &27.63 &61.55 &4.21   \\    
& DHC~\cite{wang2023dhc}$^\dagger$     &\textcolor{orange}{38.28} &\textcolor{orange}{20.34} &\textcolor{orange}{64.16} &\textcolor{orange}{3.51} \\  
& \textbf{MICD (ours)}   &\textcolor{red}{\textbf{42.75}}	&\textcolor{red}{\textbf{16.72}} &\textcolor{red}{\textbf{66.39}}	&\textcolor{red}{\textbf{3.41}}  \\

\bottomrule
\end{tabular}
}
\begin{threeparttable}
 \begin{tablenotes}
        \scriptsize
        \item[$\dagger$] implement semi-supervised segmentation methods on our dataset.
        \item[$\star$] extend semi-supervised classification methods to segmentation with CPS as the baseline.
\end{tablenotes}
\end{threeparttable}

\end{table}

\begin{table}[t] 
\centering
\setlength{\tabcolsep}{12pt}
\scriptsize
\renewcommand{\arraystretch}{0.8}
\caption{Quantitative comparison between MICD and SOTA SSL segmentation methods on \textbf{20\% and 40\% labeled SYNAPSE dataset}. \textcolor{red}{The best results are shown in red}, and \textcolor{orange}{the second-best results are shown in orange}. 
}
\label{sota4}
\resizebox{\columnwidth}{!}{ 

\begin{tabular}{c|c|c@{\ \ }c|c@{\ \ }c}
\toprule
\multicolumn{2}{c|}{}   &\multicolumn{2}{c|}{20\%}    &\multicolumn{2}{c}{40\%} \\
\multicolumn{2}{c|}{\multirow{1.5}{*}{Methods}}  &Avg. &Avg. &Avg. &Avg.    \\ 
\multicolumn{2}{c|}{}   &Dice &ASD &Dice &ASD   \\
\midrule
\multirow{11.5}{*}{\rotatebox{90}{General}}
& V-Net (fully) & 62.09±1.2 &10.28±3.9 & 62.09±1.2 &10.28±3.9 \\ 
\midrule
& UA-MT~\cite{yu2019uamt}$^\dagger$     &20.26±2.2 &71.67±7.4 &17.09±2.97 &91.86±7.93 \\
& URPC~\cite{luo2021urpc}$^\dagger$     &25.68±5.1 &72.74±15.5 &24.83±8.19 &74.44±17.01 \\
& CPS~\cite{chen2021cps}$^\dagger$      &33.55±3.7 &41.21±9.1 &33.07±1.07 &60.46±2.25 \\
& SS-Net~\cite{wu2022ssnet}$^\dagger$   &35.08±2.8 &50.81±6.5 &32.98±10.99 &71.18±20.77 \\
& DST~\cite{chen2022dst}$^\star$        &34.47±1.6 &37.69±2.9 &35.57±1.54 &55.69±1.43 \\
& DePL~\cite{wang2022depl}$^\star$      &36.27±0.9 &36.02±0.8 &36.16±2.08 &56.14±7.61 \\ 
\midrule
\multirow{8.5}{*}{\rotatebox{90}{Imbalance}}
& Adsh~\cite{guo2022adsh}$^\star$       &35.29±0.5 &39.61±4.6 &35.91±6.17 &53.7±6.95 \\  
& CReST~\cite{wei2021crest}$^\star$     &38.33±3.4 &22.85±9.0 &41.6±2.49 &27.82±5.07 \\  
& SimiS~\cite{simis}$^\star$            &40.07±0.6 &32.98±0.5 &47.09±2.33 &33.46±1.75 \\  
& Basak \textit{et al.}~\cite{basak2022addressing}$^\dagger$        &33.24±0.6 &43.78±2.5 &35.03±3.68 &60.69±6.57  \\      
& CLD~\cite{lin2022cld}$^\dagger$       &41.07±1.2 &32.15±3.3 &48.23±1.02 &28.79±3.64   \\    
& DHC~\cite{wang2023dhc}$^\dagger$     &\textcolor{orange}{48.61±0.9} &\textcolor{orange}{10.71±2.6} &\textcolor{orange}{57.13±0.8} &\textcolor{orange}{11.66±2.7} \\  
& \textbf{MICD (ours)}   &\textcolor{red}{\textbf{54.1±1.46}}	&\textcolor{red}{\textbf{9.32±1.8}} &\textcolor{red}{\textbf{65.83 ± 0.28}}	&\textcolor{red}{\textbf{5.99±1.13}}  \\

\bottomrule
\end{tabular}
}
\begin{threeparttable}
 \begin{tablenotes}
        \scriptsize
        \item[$\dagger$] implement semi-supervised segmentation methods on our dataset.
        \item[$\star$] extend semi-supervised classification methods to segmentation with CPS as the baseline.
\end{tablenotes}
\end{threeparttable}

\end{table}

\section{Experiments}
\subsection{Experimental Setup}
\subsubsection{Dataset}
Our proposed semi-supervised medical image segmentation method has been validated on the AMOS~\cite{amos} and Synapse~\cite{synapse}  datasets. Specifically, the AMOS dataset contains 360 CT scans with 15 foreground categories. In contrast, the Synapse dataset comprises 30 augmented abdominal CT scans covering 13 foreground categories. The dataset divisions follow the DHC~\cite{wang2023dhc}. To minimize the effect of randomness due to the limited sample size, we performed a triple-fold validation on the Synapse dataset using different random seeds.
\subsubsection{Implementation Details}
In this study, a CNN-based VNet~\cite{milletari2016v} is basic of networks. For both the AMOS~\cite{amos} and Synapse~\cite{synapse} datasets, we used an SGD optimizer with a momentum of 0.9, an initial learning rate of 0.01, and a ``poly'' decay strategy~\cite{isensee2021nnunet}. The batch size was 2, consisting of 1 labeled and 1 unlabeled data sample. All methods were implemented in Python 3.8.19 using PyTorch 2.2 on an Nvidia GeForce RTX 4090 GPU with CUDA 11.8.
\subsubsection{Evaluation metrics}
We utilize three key metrics: Avg. Dice, Avg. ASD, and Average Dice of Each Class. Avg. Dice measures the overall similarity between segmented and ground truth regions for segmentation accuracy. Avg. ASD quantifies the average distance between surfaces to assess boundary precision. Average Dice of Each Class enables evaluating performance of individual anatomical structures. These metrics are selected as they comprehensively evaluate our method's effectiveness in medical image segmentation from both overall and specific class aspects. 

\begin{table}[t]
\centering
\setlength{\tabcolsep}{12pt}
\scriptsize
\renewcommand{\arraystretch}{0.6} 
\caption{The ablation study of Masked Cross Pseudo Consistency (MCPC), Cross Feature Consistency (CFC) and Cross Model Discrepancy (CMD) on the \textbf{5\% labeled AMOS dataset.}}
\label{ablation}
\resizebox{\columnwidth}{!}{
\begin{tabular}{ccc|cc}
\toprule
\multicolumn{3}{c|}{{Methods}} & \multicolumn{2}{c}{{Metrics}} \\ 
\midrule
MCPC &CFC &CMD &Avg Dice. &Avg ASD. \\ 
\midrule
${\times}$ &${\times}$ &${\times}$ &49.53 &13.89 \\
\midrule
$\checkmark$ &${\times}$ &${\times}$ &53.87 &8.81 \\
${\times}$ &$\checkmark$ &${\times}$ &51.74 &9.45 \\
${\times}$ &${\times}$ &$\checkmark$ &51.77 &9.28 \\
\midrule
$\checkmark$ &$\checkmark$ &${\times}$ &54.56 &6.27 \\
$\checkmark$ &${\times}$ &$\checkmark$ &55.02 &6.09 \\
${\times}$ &$\checkmark$ &$\checkmark$ &51.92 &9.91 \\
\midrule
$\checkmark$ &$\checkmark$ &$\checkmark$ &56.11 &6.22 \\ 
\bottomrule
\end{tabular}
}

\end{table}

\subsection{Result and Analysis}
Table~\ref{sota1} and Tables~\ref{sota3}-\ref{sota4} present the quantitative outcomes on the AMOS and Synapse datasets, wherein different percentages (2\%, 5\%, 10\%, 20\%, and 40\%) of labeled data are employed during the training process. Evidently, the proposed MICD method exhibits a remarkable superiority over other state-of-the-art semi-supervised learning (SSL) segmentation approaches in terms of the average Dice score and Average Surface Distance (ASD). For instance, when compared with the previous state-of-the-art method DHC on the AMOS dataset with 5\% labeled data, MICD attains a 6.58\% improvement (56.11\% vs. 49.53\%). Analogously, on the AMOS dataset with 2\% and 10\% labeled data, MICD respectively demonstrates enhancements of 4.47\% (42.75\% vs. 38.28\%) and 2.23\% (66.39\% vs. 64.16\%) in the Dice score compared to DHC. Regarding the Synapse dataset, MICD once again outperforms DHC, achieving Dice score increments of 5.49\% (54.1\% vs. 48.61\%) with 20\% labeled data and 8.7\% (65.83\% vs. 57.13\%) with 40\% labeled data.

\begin{figure}[!ht]
    \centering
    \hspace{-0.2cm}
    \includegraphics[scale=0.3]{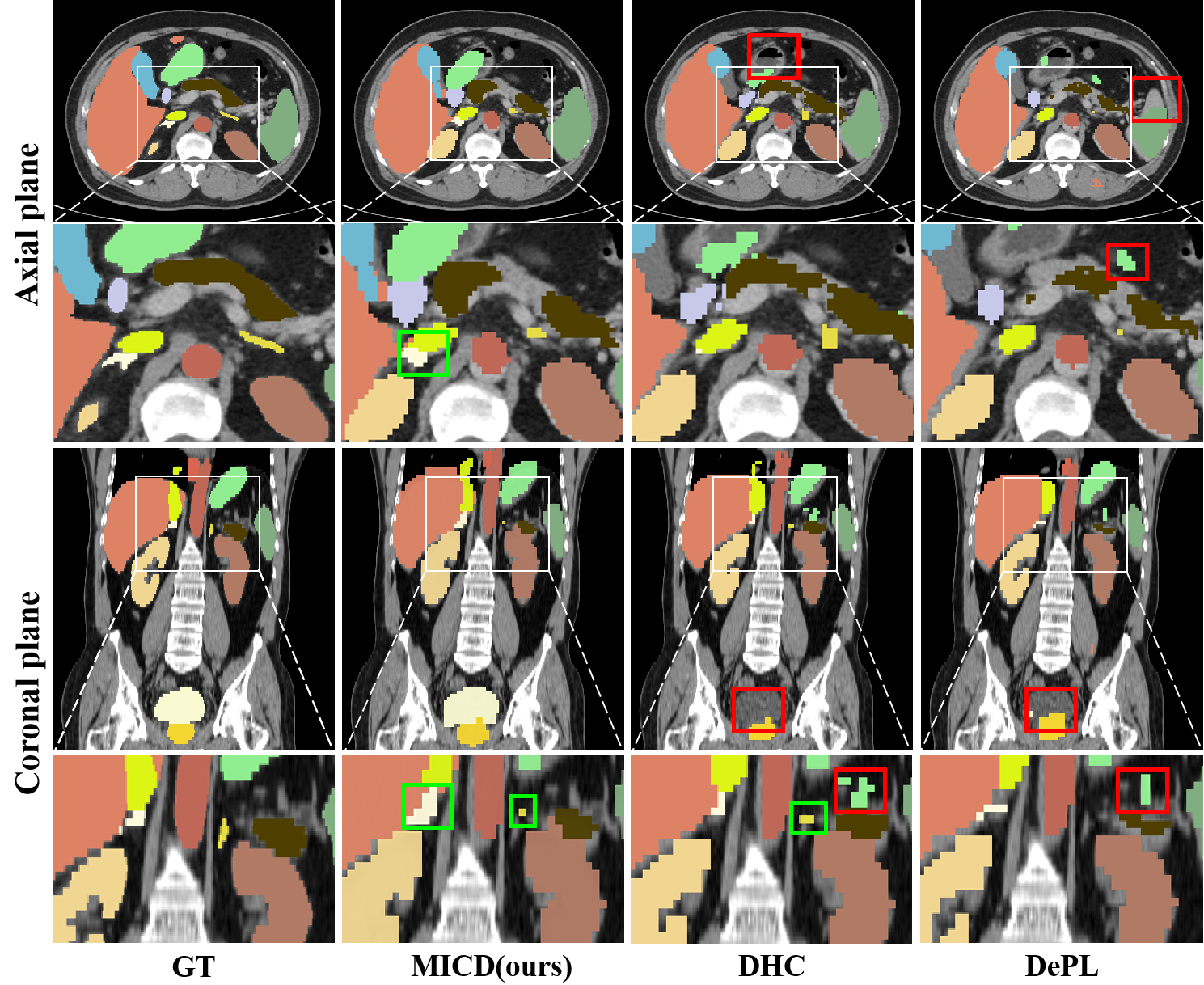}
    \vspace{-0.3cm}
    \caption{Visual comparison between MICD and the SOTA methods on 5\% labeled AMOS dataset. Regions in red boxes show poor segmentation performance, whereas those in green boxes highlight some minority classes being segmented.}
    \label{fig:vision}
\end{figure}

Furthermore, as illustrated in Table~\ref{sota1}, our method surpasses the state-of-the-art methods in nearly all classes on the AMOS dataset with 5\% labeled data, with the exceptions of the left adrenal gland (LAG), liver (Li), and prostate/uterus (P/U). Notably, it showcases outstanding performance in relatively smaller minority classes such as the gallbladder (Ga), esophagus (Es), pancreas (Pa), right adrenal gland (RAG), duodenum (Du), and bladder (Bl). This implies that our approach is effective in dealing with smaller minority classes and alleviating class imbalance to a certain degree. The visualization results in Fig.~\ref{fig:vision} provide additional corroboration. Our method demonstrates excellent performance in minority classes, as emphasized by the green boxes, and successfully averts issues of overfitting or incomplete segmentation, as denoted by the red boxes.

\subsection{Ablation Studies}
The ablation study on the 5\% labeled AMOS dataset (Table~\ref{ablation}) demonstrates the effectiveness of each component in our MICD framework, both individually and in combination. Compared to the baseline, MCPC increases the Dice score by 4.34\% (53.87\% vs. 49.53\%), highlighting its importance in capturing fine-grained local semantics and context in heterogeneous co-training. Adding CFC to MCPC further improves the Dice score by 0.69\% (54.56\% vs. 53.87\%), indicating that CFC complements MCPC by enhancing the model’s ability to learn from full data and maintain a stable spatial structure. Furthermore, the combination of MCPC and CMD further boosts the Dice score by 1.15\% (55.02\% vs. 53.87\%), demonstrating their complementary roles in enhancing robustness and accuracy. Finally, integrating all three components (MCPC, CFC, and CMD) provides the highest performance, increasing the Dice score by 6.58\% (56.11\% vs. 49.53\%), which underscores the synergistic benefits of the MICD framework in improving robustness and overall segmentation performance.

\section{Conclusion}
In this paper, we proposed the Masked Image Consistency and Discrepancy Learning (MICD) framework for semi-supervised medical image segmentation. Existing co-training methods had limitations regarding information exchange and model diversity. Our MICD framework, with its three key modules (MCPC, CFC, and CMD), overcomes these. The MCPC enriches context and small sample learning, the CFC ensures information exchange and model robustness, and the CMD preserves branch diversity. Experiments on AMOS and Synapse datasets show that our approach outperforms state-of-the-art methods. This shows that our work contributes to improving semi-supervised medical image segmentation and inspires further research in this area.

\bibliographystyle{IEEEbib}
\bibliography{icme2025references}

\end{document}